# A review of neural network algorithms and their applications in supercritical extraction

Yu Qi，Zhaolan Zheng

School of Chemical Engineering, East China University of Science and Technology，Shanghai，200237

ABSTRACT

Neural network realizes multi-parameter optimization and control by simulating certain mechanisms of the human brain. It can be used in many fields such as signal processing, intelligent driving, optimal combination, vehicle abnormality detection, and chemical process optimization control. Supercritical extraction is a new type of high-efficiency chemical separation process, which is mainly used in the separation and purification of natural substances. There are many influencing factors. The neural network model can quickly optimize the process parameters and predict the experimental results under different process conditions. It is helpful to understand the inner law of the experiment and determine the optimal experimental conditions. This paper briefly describes the basic concepts and research progress of neural networks and supercritical extraction, and summarizes the application of neural network algorithms in supercritical extraction, aiming to provide reference for the development and innovation of industry technology.

## 1. Introduction

An artificial neural network is a complex network structure formed by a large number of neurons connected to each other, which realizes certain applications by simulating certain mechanisms of the brain. Artificial neural networks used for process prediction and modeling include many models. Among them, the most widely used are feedback neural network (BPNN) and radial basis neural network (RBF), which are characterized by autonomous learning, associative storage, and the ability of rapid search the optimal solution. Therefore it is widely used in many fields such as signal processing [错误!未找到引用源。], artificial intelligence recognition[0], vehicle tracking[0], stock market index prediction [0], medical diagnosis[0], speech processing[0], as well as in the supercritical extraction process. In terms of multi-parameter optimization[0] and data prediction, the application prospects are also very broad.

Supercritical extraction technology uses supercritical fluid as the extractant to extract the target components from the liquid or solid mixture to achieve the purpose of separation. Its advantages are fast extraction speed, high efficiency and environmental protection. It is often used in essential oils[0], extraction of Chinese medicine components [0], oils and fats[10, 0]. However, in the process of supercritical extraction technology, there are many experimental variables and parameters, and it is difficult to predict the influence of various factors and the interaction between various variables. It is difficult to achieve reasonable adjustments to process conditions by using mechanism modeling, so it is necessary to use The artificial neural network model performs parameter optimization and data prediction [0]. This is because the neural network algorithm does not need to understand the process mechanism in detail. It is only on the basis of input and output data that the training and learning of input and output data can establish a supercritical Process model of the extraction process.

Many scholars have studied the application of neural networks in the supercritical extraction process, put forward various suitable neural network structure models, and optimized the learning algorithm and hidden layer structure to make the predicted results basically match the actual results. This paper briefly describes the working principle and application of neural network algorithm and supercritical extraction technology, and systematically summarizes the application research progress of artificial neural network algorithm in supercritical extraction process.

## 2. Research progress of neural network

### 2.1 Introduction to artificial neural networks

Artificial neural network is composed of three parts: input layer, hidden layer and output layer. The input variables include training samples and prediction samples; the hidden layer is the part that connects the input layer and the output layer and cannot be observed from outside. When the training set is determined, the number of nodes in the input layer and output layer is also determined. The number of output neurons should be directly related to the type of work to be performed by the neural network. There is no fixed selection criterion for the number of neurons in the hidden layer and the number of hidden layers, and there is no definite selection range. You can set the number of different hidden layers and the number of neurons, and perform training separately, and the most suitable number of layers and neurons can be determined by the training results[0].

A neuron is the smallest unit that constitutes a neural network, also called a perceptron. The composition of the perceptron includes three parts: input weight, activation function and output. By substituting the output, weight and bias of this layer of neurons into the activation function, the input of the next layer of neurons can be obtained, and so on, and finally the output of the output layer of neurons can be obtained[0].

The number of elements in the input vector is equal to the number of neurons in the input layer. The calculation process is to calculate the output vector through a series of functions, where the number of elements in the output vector is equal to the number of neurons in the output layer. By adjusting the relationship between a large number of internal nodes, the purpose of information processing is achieved. The experimental data is obtained through experiments in advance and used as input and output data to analyze the internal laws between the two, and finally form a complex nonlinear system function through these laws.

Each input has a corresponding weight value, and the output is calculated by the weighted sum of the weight and the input [0]. The specific calculation method is related to the type of neural network used. Two commonly used neural network algorithms will be introduced below: BPNN and RBF.

**BPNN** The basic step of BP neural network model establishment is to provide learning samples to the neural network, including input values and expected output values, and modify the weights according to the error between the actual output value and the expected output value, so that the modified network has the output value is as close as possible to the expected output value. The specific steps are as follows [0]:

First, take the error sum of squares of the neurons in the output layer as the objective function, as shown in equation (1), find the weight and bias when the objective function reaches the minimum value.

$$J_d = \sum_{i \in \text{output layer elements}} \left( \mathrm{t}_i - y_i \right)^2 \tag{1}$$

Among them:

$J_d$ is the error of sample d， $\mathrm{t}_i$ is the predicted value， $y_i$ is the actual value.

Next, use the stochastic gradient descent algorithm [0] (see equation (2)) to optimize the error:

$$\text{The original } \omega_{ji} - \eta \frac{\partial J_d}{\partial \omega_{ji}} = \text{The new } \omega_{ji} \tag{2}$$

Regarding the partial derivative of $J_d$ to $\omega_{ji}$, it is necessary to use the principle of the chain derivation rule for calculation, and treat it differently according to the location of the j node (hidden layer or output layer).

According to the analysis principle of backpropagation, it can be known that if you want to find the error term of node j, you must first know the error term and weight of the next layer node. Therefore, the calculation of the error term needs to start from the output layer, and then calculate the hidden layer and Enter the error term of the layer and update all the weights[0]. The weight is adjusted every time a sample is processed. After multiple rounds of iteration (that is, all the training data is processed repeatedly for multiple rounds), the weights of the network model can be trained to achieve the objective function.

**RBF** The output value of the RBF neural network is no longer 0 or 1, but a set of smooth numbers that have the largest output value at a specific input value. It is the same multi-layer forward network as the BP network, but the transfer function is different. BP usually uses the Sigmoid function or linear function as the transfer function, while the RBF network uses the radial basis function as the transfer function. Compared with BP neural network, there is only one hidden layer, so its calculation speed is much faster than BP neural network. BP neural network is a kind of global approximation neural network. When the output value changes, each input weight needs to be changed, which makes the learning speed of the network very slow. RBF neural network is a kind of local approximation neural network, so RBF neural network is better than BP neural network in function approximation, classification ability and learning speed.

### 2.2 Application of neural network

In order to effectively detect bad behavior data, Ghaleb *et al.*[0].introduced new features to represent bad behavior, environment and communication status. Using artificial neural network technology, namely feedforward and backpropagation algorithms, based on historical data containing attackers and normal traffic data, an effective error behavior classifier is trained.

Zhang et al.[0]applied the deep neural network algorithm to the artificial intelligence system, and through supervised learning, reduced the error of each training, improved the efficiency and computational stability of the neural network algorithm, and promoted the face through the optimization of the algorithm. Breakthrough in the recognition of new business functions, with a face recognition rate of 99.5%.

In addition to classification and recognition, Gao *et al.*[0] applied neural network models to generate new data. They proposed a game theoretical deep learning framework to minimize the dissimilarity between distribution of real data and the generator data. The distance between those two distributions is measured by f-divergence and optimized by solving a distributional robust game[0]. When the algorithm converges, human testers cannot distinguish the computer-generated images from the real ones.

### 2.2.3 Parameter optimization

Deng et al.[0].used the neural network platform in the JMP 7.0 software to establish a neural network model for supercritical carbon dioxide extraction of papaya seed oil, and optimized the process parameters for extracting papaya seed oil, so that the oil extraction rate reached more than 30%. Neural networks can also be used for parameter optimization of reactive distillation, extractive distillation and other processes[0].

## 3. Research progress of supercritical extraction

### 3.1 Supercritical extraction

On the P-T diagram (constant volume diagram), the fluid whose temperature and pressure are slightly higher than the critical point is called supercritical fluid[0]. The density of supercritical fluids is very sensitive to changes in temperature and pressure. A small change in pressure may make a big change in the density of the fluid and its solubility[0]. The supercritical fluid is used as the extractant to extract the substance from the liquid or solid mixture to achieve the purpose of separation. Through the control of temperature and pressure, the density of the supercritical fluid is changed to change its solubility, and then different components are selectively extracted in turn. In different pressure ranges, the obtained extracts are also different. Then, by means of decompression and temperature increase, the supercritical fluid becomes a gas, which is separated from the extracted substance [0].

Several compounds have been proven to be supercritical fluid extraction solvents. For example, hydrocarbons such as chloroform, n-butane, nitrous oxide, sulfur hexafluoride and fluorinated hydrocarbons [0]. However, carbon dioxide ($CO_2$) is the most popular supercritical fluid extraction solvent because $CO_2$ is harmless to human health and the environment, and it complies with sustainability standards; secondly, its critical temperature (31.2 ℃) is the biological activity in the preserved extract. The key problem of the compound [0]; finally, the extract will not come into contact with the air, because the air may undergo a slight oxidation reaction. Since carbon dioxide is a gas at room temperature, when the extraction is completed and the system is depressurized, the carbon dioxide is eliminated and a solvent-free extract is obtained. On an industrial scale, when carbon dioxide consumption is high, operations can be controlled to recover carbon dioxide. However, due to its lower polarity [0], $CO_2$ is less efficient in extracting more polar compounds from natural matrices. Modifiers (also called co-solvents) are usually used to overcome this problem [0].

### 3.2 Application of supercritical extraction

Yousefi *et al.* [0] compared supercritical extraction technology with other methods of extracting essential oils and found that supercritical extraction can extract high-quality essential oils at a relatively mild temperature and within a short period of time, with minimal loss of components. Therefore, it is suitable to use supercritical extraction technology to extract essential oils from plants.

Molino *et al.* [0] used supercritical $CO_2$ technology to investigate the effects of time,

temperature and pressure on the extraction of astaxanthin and lutein from Haematococcus pluvialis. Experimental results show that temperature plays a vital role in the extraction of astaxanthin and lutein. At 65 °C and 550 bar, and the extraction rates of the two are the highest, 92% and 93%, respectively.

Lu et al. [0] used supercritical $CO_2$ extraction technology to extract camellia seed oil, and used BP neural network to simulate and predict the process, and the model prediction results obtained were good. Deng *et al.* [0] compared the extraction effects of supercritical $CO_2$ extraction and solvent extraction on papaya seed oil, and found that the solvent extraction method has problems such as low oil yield, high impurity content, and solvent residues. The supercritical $CO_2$ extraction method not only has a high extraction rate, but also all physical and chemical properties meet the requirements. Tang *et al.* [0] used supercritical extraction technology to extract nutmeg oil, and adopted an orthogonal experiment scheme to obtain suitable process conditions and the highest oil yield is 45.6%.

Supercritical extraction technology can be used for the extraction of traditional Chinese medicine elements, including coumarin, cordycepin and so on. Song *et al.* [错误!未找到引用源。] and Liu *et al.* [错误!未找到引用源。] used supercritical extraction technology to extract cordycepin from artificial pupa seeds and extract pre-oxidation and imperatorin from Angelica dahurica, respectively.

**4. Neural network research on supercritical extraction process**

The supercritical extraction process can be applied to the extraction of various substances such as natural substances and Chinese medicine elements. Using neural networks, by continuously inputting experimental data into the network for training, the most accurate model can be obtained to optimize the process parameters of the supercritical extraction process and predict the experimental results.

Lu *et al.* [0] used BP neural network to simulate and predict the process based on the experimental data of supercritical CO2 extraction of camellia seed oil. Using camellia seeds (the oil content of the raw material measured by Soxhlet extraction method was 41.11%) as the raw material, the effects of average particle size, CO2 flow rate, extraction pressure, extraction temperature and extraction time on the extraction rate were investigated. In order to improve the learning speed and convergence of the BP network, a three-layer BPNN with 5 and 10 hidden neurons was constructed, and the training samples obtained from orthogonal experiments were used to investigate the adaptive learning rate method (BPA) and additional momentum Method (BPM), additional momentum and adaptive adjustment of learning rate combined algorithm (BPX) and Levenberg-Marquart algorithm (LM) influence on BP network training. From the verification and test results, it can be seen that the over-computed mean square error (MSE) of the BP algorithm is significantly higher than that of the LM algorithm, and the correlation coefficient (R) of the BPA and BPX algorithms is lower than that of the LM algorithm, indicating that the training accuracy of the LM algorithm is high. The generalization ability of the network is good, and the training effect is better than BP, BPA and BPX algorithms. Therefore, L-M algorithm is a suitable learning algorithm for BPNN for supercritical $CO_2$ extraction of camellia seed oil. The appropriate number of hidden layers and the number of neurons can improve the accuracy of network training, so the hidden layer structure needs to be optimized. By comparing the properties of single hidden layers with different numbers of neurons and investigating the performance of double hidden layers, it is found that the neural network with a structure of 5/8/1 has the best simulation performance, so this model is used as the neural network model. The experimental data of supercritical $CO_2$ extraction of camellia seed oil are used as training samples and prediction samples, the tan-sigmoid function is used as the transfer function for this time, and the L-M algorithm is the learning algorithm to construct a neural network model for the supercritical $CO_2$ extraction process. Using the obtained neural network model to predict the experimental results, it is found that the predicted value has a larger error compared with the experimental value when the extraction is 30 minutes, and the error of the other predicted value and the experimental value is within 3%, which is in good agreement. The reason for the large error at 30 minutes of extraction may be that the extraction process is easy to fluctuate at the beginning of the experiment, which leads to deviations in the experimental value.

Deng *et al.* [0] used supercritical CO2 extraction technology to extract papaya seed oil. The data was established by Box-Behnken response surface test design, and the neural network model of supercritical CO2 extraction of papaya seed oil was established with the aid of the neural network platform in the JMP 7.0 software, and the technological parameters of the extraction of papaya seed oil were optimized and determined. First, the optimal material particle size and $CO_2$ flow rate are obtained through experiments using the fixed variable method, which are 20 mesh and 25 L/h. Then a 3/3/1 neural network model was established, in which the three input variables were extraction pressure, temperature and time, and the output variable was the extraction rate of papaya seeds. Through continuous training, a suitable neural network model is obtained. Finally, the neural network is tested, and the results show that the experimental values are in good agreement with the predicted values. The established neural network model is accurate and stable, so it can be used to predict the supercritical CO2 extraction process of papaya seed oil. According to the relationship between the oil yield and the extraction pressure, temperature and time obtained by the neural network, the optimal operating conditions are extraction pressure of 27 MPa, temperature of 54°C, and time of 3 h. Under these conditions, the extraction yield can reach more than 30%.

Tang *et al.* [0] combined the genetic algorithm with the BP algorithm to form a genetic BP neural network, which was applied to the process of supercritical extraction of nutmeg oleoresin and found that the network model has a higher accuracy rate. By applying the genetic BP neural network to the orthogonal experimental design, more reasonable structural parameters can be directly obtained, and the training effect of the training samples can be enhanced, and the mathematical model can be automatically obtained from the experimental data. First, the orthogonal experiment scheme is adopted to obtain 25 experimental data, and the best scheme in the experiment can be obtained by comparison (extraction pressure 25 MPa, CO2 pump frequency 20 Hz, extraction temperature 45℃, extraction time 2 h, separation pressure 7 MPa, separation temperature 20℃), under these conditions, the oil yield is 45.6%. Based on experience, a neural network model with 15 neurons in the hidden layer is used in the extraction process of nutmeg oleoresin. The training model was used to test the training effect of the sample, and it was found that the average error of the oil product rate was 1%, indicating that the neural network has a good simulation effect and can be used to optimize the parameters of the nutmeg oleoresin. Under optimized conditions (extraction pressure 27 MPa, carbon dioxide pump frequency 18 Hz, extraction temperature 47°C, extraction time 2 h, separation pressure 8 MPa, separation temperature 20°C), the extraction rate is 46.12%, which is 0.52% higher than the highest yield obtained by orthogonal experiment.

Liu *et al.* [错误!未找到引用源。] took the sum of the yields of imperatorin and oxidized anthurium (the effective component of Angelica dahurica) as the objective function, and predicted the supercritical extraction results of Angelica dahurica with a structure of 5/4/3 network, where the input variables are extraction pressure, extraction temperature, extraction time, separation pressure and the granularity of the medicinal materials, and the output variables are the yield of total coumarin, oxidized prehusin and imperatorin. 18 sets of orthogonal experimental data are used as training samples, and the trainbr function provided by Matlab 6.5 neural network toolbox is used to train the network, and then four sets of data are randomly generated within the experimental operating range for experiments, and the results obtained are used as test samples to test the trained network. After testing, it is found that the test results are in good agreement with the actual values, and the relative error is less than 5%.

Since understanding the effects of extraction temperature, extraction pressure, extraction time, and amount of entrainer on the extraction rate of cordycepin is the basis for the supercritical CO2 extraction process, it is difficult to predict the effects of changes in various factors using supercritical extraction technology, so Song *et al.*[错误!未找到引用源。] .established a BP neural network model for the supercritical extraction of cordycepin to investigate the interaction of multiple process parameters on the extraction results. First, determine the number of hidden layer nodes. There is currently no uniform standard for determining the number of hidden layer nodes. In this simulation, the principle of minimum square error is used to determine the number of hidden layer nodes. By analyzing the error curve of different hidden layer node numbers, it can be known that when the number of nodes is 17, the training error of the sample is the smallest, so the number of hidden layer nodes is determined to be 17. After that, an artificial neural network with a 5/17/1 structure was established for training and prediction, and the simulation effect of the neural network model obtained was very good. The calculation results are in accordance with the experimental rules, indicating that the neural network simulation of the supercritical extraction process is feasible, and provides a theoretical basis for optimizing the parameters of the extraction process.

## 5. Conclusions

The artificial neural network includes three layers, input layer, output layer and hidden layer. Neurons are the smallest unit that constitutes the artificial neural network. The neurons of each layer are connected through activation functions, weights and biases. By inputting the training data into the neural network, the weight value is continuously modified to achieve the objective function value. Because neural networks have the advantages of autonomous learning, associative storage and rapid search for optimal solutions, they can be applied in many fields, such as intelligent driving, signal processing, and multi-parameter optimization calculations. It can also be used to simulate various complex chemical processes. Optimization and control.

Supercritical extraction has the advantages of high extraction rate, fast extraction speed, moderate extraction temperature, and no solvent residues, so it can be used in many fields, including the extraction of essential oils, carotenoids, and oils. There are many experimental factors that affect the supercritical extraction experiment. It is difficult to realize the interaction between the various factors and the effect on the experimental results only by experimental research. Therefore, it is necessary to use neural network for parameter optimization and data prediction to find the optimal Process conditions, save costs.

This article reviews some scholars using several neural network algorithms including BP neural network to predict and optimize the supercritical extraction process of essential oils, carotenoids, and lipids, and obtain the prediction results and actual results of the neural network model. The difference is small and can be used for parameter optimization and data prediction. This shows that the research of supercritical extraction process with the help of artificial neural network model has great application prospects, but people still need to continuously improve the neural network model, optimize the training algorithm, and improve the training speed and accuracy of the results